\RequirePackage{silence}
\documentclass[pdflatex,sn-mathphys-num]{sn-jnl}

\usepackage{silence}
\usepackage{newtxtext}
\usepackage{newtxmath}

\usepackage{textcomp}
\usepackage{gensymb}

\usepackage{bibunits}
\usepackage{natbib}

\setcitestyle{super,open=,close=}

\defaultbibliography{sn-bibliography}
\defaultbibliographystyle{sn-mathphys-num}

\makeatletter
\newcommand*{\newbibstartnumber}[1]{
  \apptocmd{\thebibliography}{
    \global\c@NAT@ctr #1\relax
    \addtocounter{NAT@ctr}{-1}
  }{}{}
}
\makeatother

\usepackage{lineno}
\nolinenumbers
\usepackage{amsmath}

\usepackage{booktabs}
\usepackage{makecell}
\usepackage{multirow}

\usepackage{subcaption}
\graphicspath{{figures}}

\usepackage{enumitem}
\setlist{topsep=.5em}
\setlist[description]{leftmargin=0.5cm,labelindent=0cm}

\begin{document}

\title[GlobalBuildingMap]{GlobalBuildingMap --- Unveiling the Mystery of Global Buildings}

\author*[1,2]{\fnm{Xiao Xiang} \sur{Zhu}}\email{xiaoxiang.zhu@tum.de}
\author[1]{\fnm{Qingyu} \sur{Li}}
\author[3]{\fnm{Yilei} \sur{Shi}}
\author[1]{\fnm{Yuanyuan} \sur{Wang}}
\author[1]{\fnm{Adam J.} \sur{Stewart}}
\author[1]{\fnm{Jonathan} \sur{Prexl}}

\affil[1]{\orgdiv{Chair of Data Science in Earth Observation}, \orgname{Technical University of Munich}, \orgaddress{\street{Arcisstraße 21}, \postcode{80333} \city{Munich}, \country{Germany}}}
\affil[2]{ \orgname{Munich Center for Machine Learning}, \orgaddress{\postcode{80333} \city{Munich}, \country{Germany}}}
\affil[3]{\orgdiv{School of Engineering and Design}, \orgname{Technical University of Munich}, \orgaddress{\street{Arcisstraße 21}, \postcode{80333} \city{Munich}, \country{Germany}}}

\abstract{
Understanding how buildings are distributed globally is crucial to revealing the human footprint on our home planet. This built environment affects local climate, land surface albedo, resource distribution, and many other key factors that influence well-being and human health. Despite this, quantitative and comprehensive data on the distribution and properties of buildings worldwide is lacking. To this end, by using a big data analytics approach and nearly 800,000 satellite images, we generated the highest resolution and highest accuracy building map ever created: the GlobalBuildingMap (GBM). A joint analysis of building maps and solar potentials indicates that rooftop solar energy can supply the global energy consumption need at a reasonable cost. Specifically, if solar panels were placed on the roofs of all buildings, they could supply 1.1--3.3 times --- depending on the efficiency of the solar device --- the global energy consumption in 2020, which is the year with the highest consumption on record. We also identified a clear geospatial correlation between building areas and key socioeconomic variables, which indicates our global building map can serve as an important input to modeling global socioeconomic needs and drivers.
}

\keywords{Earth observation, global buildings, big data analytics, solar potential, socioeconomic analysis}

\maketitle

\begin{bibunit}

Identifying all buildings on the planet would further a better understanding of human activities across the globe, which is crucial for making key strategic decisions needed to tackle the grand societal challenges of the century, such as urbanization and climate change, and further analyze their impacts. For example, more than half the population of the planet (currently about 56\%) live in urban areas~\cite{united_nations_world_2019}. This built environment influences local and regional weather and climate, water runoff characteristics and flood risk, energy efficiency, resource allocation and distribution, and numerous other factors that affect well-being. The characteristics of the built environment such as building density, green spaces, traffic flow, and so on have a profound impact, therefore, on some four billion people. Robust and comprehensive information on these characteristics and how they change with time is essential for urban planning and modeling impacts.

In the context of climate change, the latest 2022 IPCC assessment report (AR6)~\cite{ipcc_summary_2022} shows that global net anthropogenic greenhouse gas (GHG) emissions in 2019 reached about 60~GtCO\textsubscript{2}-eq, 12\% higher than that in 2010, which is the highest increase in average decadal emissions on record. Buildings contributed 21\% of the GHG emissions in 2019. The expansion of human settlements naturally increases GHG emissions, due to the increase of population and production activities, as well as the loss of habitat, biomass, and carbon storage~\cite{seto_global_2012}. On the other hand, compact urban structures can also positively contribute to reducing GHG emissions~\cite{glaeser_triumph_2011}. IPCC AR6 included a separate chapter on buildings and suggested that new buildings and existing ones, if retrofitted, are projected to approach net zero GHG emissions in 2050 if ambitious renewable energy measures are implemented~\cite{ipcc_summary_2022}. Therefore, applying a globally consistent mapping approach to understand the human settlement expansion can significantly contribute to the 2050 net zero carbon goal. 

\begin{table}[htbp]
    \centering
    \caption{Comparison between different building products. GBM has the highest resolution and highest accuracy of any complete global building map.}
    \label{tab:products}
    \begin{tabular}{@{}ccccc@{}}
        \toprule
        \textbf{Product} & \textbf{Spatial coverage} & \textbf{Completeness} & \textbf{Temporal coverage} & \textbf{Spatial resolution} \\
        \midrule
        GUF~\cite{esch2017breaking} & Global & Complete & 2011 & 12~m \\
        HRSL~\cite{facebook2016hrsl} & Global & Complete & 2015 & 30~m \\
        GHSL~\cite{pesaresi2023ghs} & Global & Complete & 2018 & 10~m \\
        WSF~\cite{marconcini2021understanding} & Global & Complete & 2019 & 10~m \\
        WSF 3D~\cite{esch2022world} & Global & Complete & 2011--2013 & 90~m \\
        \midrule
        OSM~\cite{osm2022planet} & Global in part & Incomplete & Unknown & Unknown \\
        Google~\cite{sirko2021continental} & \makecell{Africa, South and Southeast Asia, \\ Latin America,  Caribbean} & Complete & Unknown & 0.5~m \\
        Microsoft~\cite{microsoft2023building} & Global in part except for China & Incomplete & 2014--2023 & 0.3--0.6~m \\
        \midrule
        GBM & Global & Complete & 2018--2019 & 3~m \\
        \bottomrule
    \end{tabular}
\end{table}

Given the urgent need, a number of building datasets have been created, as documented in Table~\ref{tab:products}. To date, the only global datasets include the Global Urban Footprint (GUF)~\cite{esch2017breaking}, High-Resolution Settlement Layer (HRSL)~\cite{facebook2016hrsl}, Global Human Settlement Layer (GHSL)~\cite{pesaresi2023ghs}, and World Settlement Footprint (WSF)~\cite{marconcini2021understanding}. However, all of these raster datasets lack the spatial resolution necessary to identify smaller buildings or to generate vector data for individual buildings. Even though buildings only occupy a relatively small fraction of the land surface, they drive global environmental changes. Understanding the expansion pattern both broadly and down to the detail of individual buildings enables unprecedented applications. For example, the literature shows that compact urban structures contribute to reducing greenhouse gas emissions~\cite{glaeser_triumph_2011}, but could also worsen the urban environment through the urban heat island effect. A global building map also lays the foundation for creating global 3D building models for better study of the above-mentioned points, as multiple papers demonstrate the importance of including a vertical dimension in the building map~\cite{li_continental-scale_2020,esch2022world}. We believe the location and extent of individual buildings are crucial to revealing the true footprint of humankind.

Current efforts toward mapping individual buildings on a large scale are led, for example, by OpenStreetMap (OSM)~\cite{osm2022planet}, Microsoft~\cite{microsoft2023building}, and Google~\cite{sirko2021continental}. However, there is a large gap between tailored local-/regional-scale mapping and systematic global-scale mapping of buildings. Both the completeness and coverage of the existing building maps are limited. OSM is the largest open database of building maps, but mainly covers Europe and part of North America. The coverage in Asia, Africa, South America, and Oceania is opportunistic since OSM includes only voluntarily contributed geographic information. This also leads to inconsistent quality across different countries and regions~\cite{herfort2023spatio}. The Microsoft Building Footprint dataset covers most of the globe except for China, but is largely incomplete. The Google building footprint covers Africa, South Asia, Southeast Asia, Latin America, and the Caribbean. As of April 2024, OSM contains 600M buildings, Microsoft has mapped 1.3B buildings, and Google has mapped 1.8B buildings. All of these are a small subset of the 4.1B buildings estimated by the UN's 2019 Urban Indicators Database~\cite{un2019habitat}. None of the above-mentioned endeavors alone or their combinations offers global-scale building maps.

\section*{GlobalBuildingMap}

The availability of high-resolution, near global coverage, satellite imagery, as well as recent technological advancements in data science present the opportunity to close this knowledge gap. Here, we present the GlobalBuildingMap (GBM): the highest accuracy and highest resolution global building map ever created. GBM is derived from nearly 800,000 PlanetScope satellite images, and is distributed in the form of a binary raster (building and non-building) at a resolution of 3 meters. Such high spatial resolution is crucial for detecting smaller buildings and temporary shelters.

\begin{figure}[htbp]
    \centering    
    \includegraphics[width=\textwidth]{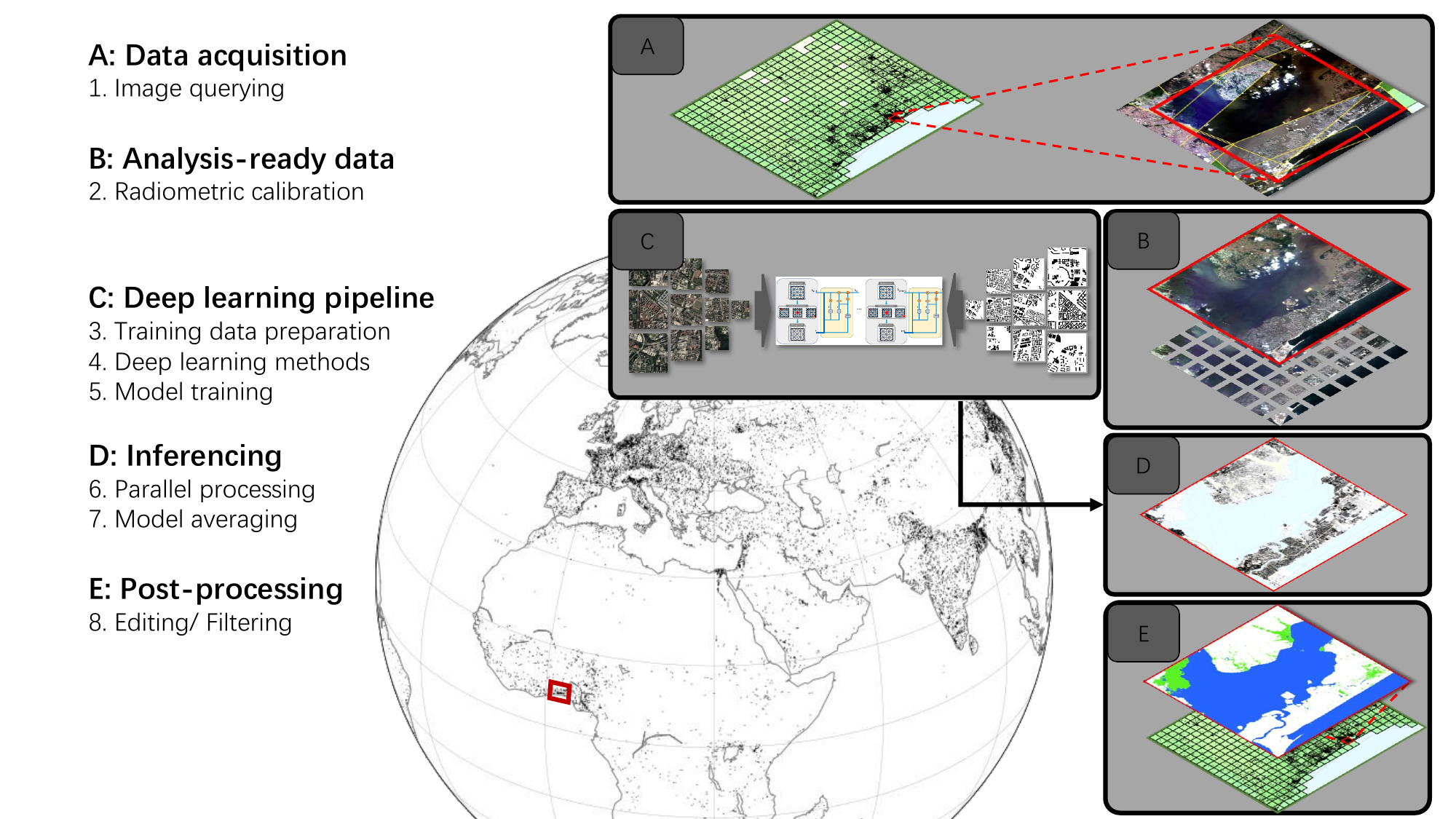}
    \caption{The big data analytics approach used to generate building footprints at a global scale. The workflow can be briefly summarized into five steps: data acquisition, analysis-ready data preparation, machine learning, inferencing, and post-processing. We use the Global Urban Footprint (GUF) to detect built-up areas on a 0.2\degree{}~\(\times\)~0.2\degree{} grid. Planet APIs were employed to acquire either Surface Reflectance or Basemap images based on the cloud coverage. Those products were calibrated and mosaicked for each 0.2\degree{}~\(\times\)~0.2\degree{} urban cell. The inferencing step takes those mosaics and four pretrained models as input, and produces the raw global building maps. Post-processing includes filtering false positives in non-urban areas with land cover layers and visualization.}
    \label{fig:workflow}
\end{figure}

The automatic building segmentation framework shown in Fig.~\ref{fig:workflow} is based on deep learning techniques. We collected over 100,000 pairs of OSM building footprint masks and PlanetScope satellite data, each with a size of 256~\(\times\)~256~pixels, selected from 74 cities across the globe, paying special attention to the areas with scarce geoinformation, such as Africa and South America. Among them, 20\% were selected as validation samples to evaluate the performance of the trained models. We trained four state-of-the-art deep learning models~\cite{shi2020building,jegou2017one,baheti2020eff} with the prepared training dataset and fused their results by majority voting: a building pixel is determined if at least two of the four models predict it to be a building pixel. The trained models were employed for the inference of the global PlanetScope data by means of a complete pipeline of data download, analysis-ready data preparation, inferencing, and post-processing. We used the Global Urban Footprint (GUF)~\cite{esch2017breaking} to detect settlement areas on a 0.2\textdegree{}~\(\times\)~0.2\textdegree{} grid. However, we posit that our building map does not depend on the quality of the GUF, because even if only a single pixel of a grid cell appears as built in GUF, we will process the entire grid cell with our pipeline. Planet APIs were employed to acquire either Surface Reflectance or Basemap images, depending on the cloud coverage. Those images were calibrated and mosaicked for each grid cell. The inferencing step then took those mosaics and four pretrained  models  as  input,  and  produced  the  raw  global  building footprints. As the last step, the raw global building footprints were filtered by removing false alarms using land cover layers, including WSF~\cite{marconcini2020outlining} and FROM-GLC10~\cite{chen2019stable}.

In total, the global building areas shown in the GBM amount to 0.67~million~km\textsuperscript{2}, 2.35 times more than the 0.2~million~km\textsuperscript{2} estimated by \citet{joshi2021}. The latter estimate was made by means of a machine learning-based regression model trained using regional reference data collected from different data sources. This means, different from our approach which segments out individual buildings with a spatial resolution of 3~m and then sums building area, building areas are directly estimated at a resolution of 10~km. Besides the limited accuracy and generalizability of regression models, the underestimation of building areas from \citet{joshi2021} could also be associated with a bias in the training data: buildings that were constructed after the year 2015 were discarded in the training data, and the reference data are not representative enough, as regions with sparse rooftops were not considered during model training. This confirms the importance of the more precise approach used in this study, in which individual buildings are detected, to avoid possible bias.

\begin{figure}[htbp]
    \centering
    \includegraphics[width=\textwidth]{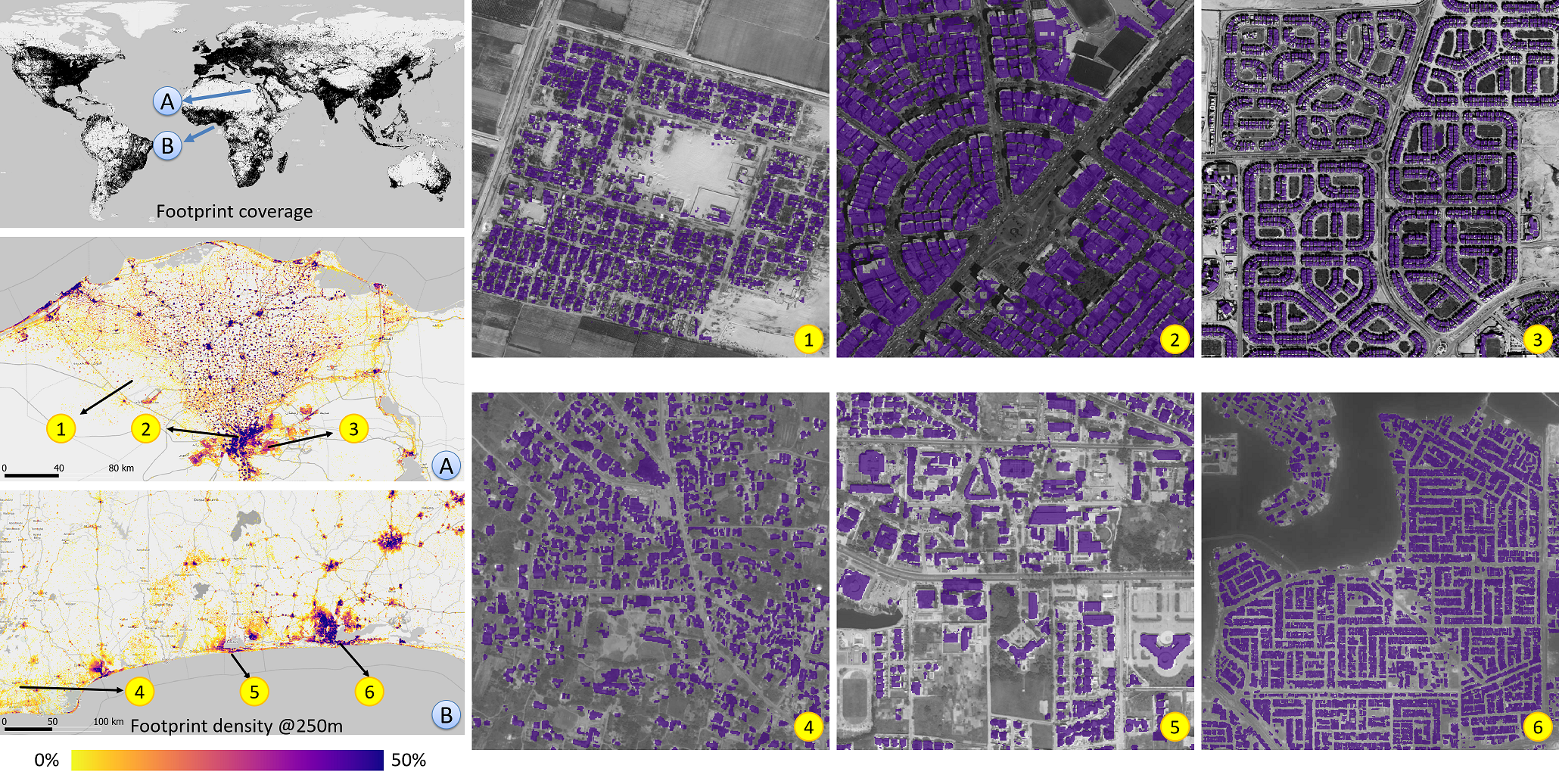}
    \caption{A glance at the GlobalBuildingMap. The black area in the upper left subfigure shows the coverage of the GBM. Two areas, (A) Cairo, Egypt and (B) Lagos, Nigeria, are exemplified by showing their building footprint density as a percentage in a 250~m~\(\times\)~250~m patch, and their detailed individual building footprints at the city block level. It demonstrates that GBM not only shows the clustering of human settlements, but also clearly indicates individual buildings.}
    \label{fig:glance}
\end{figure}

Fig.~\ref{fig:glance} shows the coverage of the GBM and a few examples of the building density and building footprint in Africa, where detailed building information is scarce. Locations (A) and (B) illustrate the building density of areas around Cairo, Egypt and Lagos, Nigeria, derived from the GBM by counting the percentage of 3~m~\(\times\)~3~m building pixels within every 250~m~\(\times\)~250~m patch. Different urban morphologies can be observed for those two urban agglomerations, as seen in locations (1--6). It also demonstrates that the GBM can provide detailed information at the individual building level.

\begin{figure}[htbp]
    \centering
    \includegraphics[width=\textwidth]{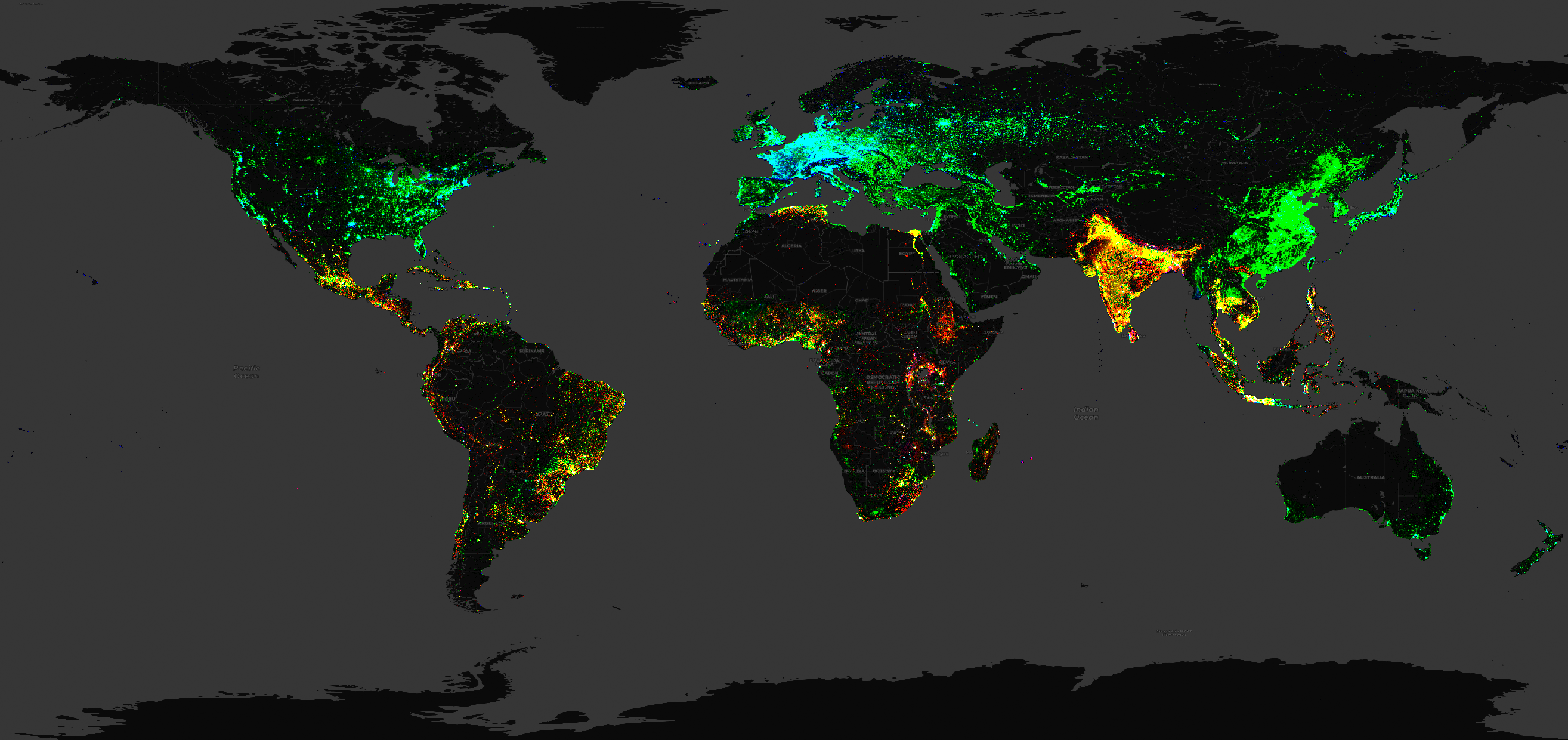}
    \caption{Global building density from three sources: our GBM (green), Google (red), and OSM (blue). Pure RGB color indicates only one source is available. Blended colors, including cyan, yellow, magenta, and white, indicate multiple sources are available. It can be seen that our footprint is the only source covering the entire globe. Google is only available in southern regions, and has a slight advantage in a few spots in eastern Africa and India, as red dominates those regions. OSM aligns well with our building footprint in Europe, as well as some Asian and North American cities, but is not available elsewhere. Only GBM covers some of the most populous regions in the world in East Asia.}
    \label{fig:density}
\end{figure}

Fig.~\ref{fig:density} draws a clear picture of the coverage of the building footprint provided by Google (red), GBM (green), and OSM (blue). A pure red, green, or blue color indicates that only one building data source out of the three is available, namely Google, GBM (ours), or OSM, respectively. Blended colors, including cyan, yellow, magenta, white, and those in between indicate that multiple data sources are available. In particular, cyan means that both OSM and GBM are available, yellow means that both Google and GBM are available, magenta shows that only Google and OSM are available, while white indicates that all data sources are available. It is apparent that GBM is the only source that has global coverage. Google is only available in southern regions and has a certain advantage in some spots (red in color) of eastern Africa and India. OSM matches well with our GBM in Europe, as well as some Asian and North American cities, but it is generally not available elsewhere. Among these data sources, the most populous regions in the world in East Asia are only covered by the GBM.

\begin{figure*}[htbp]
    \centering
    \setlength\tabcolsep{1.75pt}
    \begin{tabular}{ccccc}
        ~ & Cairo, Egypt & \shortstack{São José dos \\ Campos, Brazil} & Poitiers, France & Chengdu, China \\
        \rotatebox[origin=l,y=3mm]{90}{OSM} & 
        \frame{\includegraphics[height=.125\linewidth]{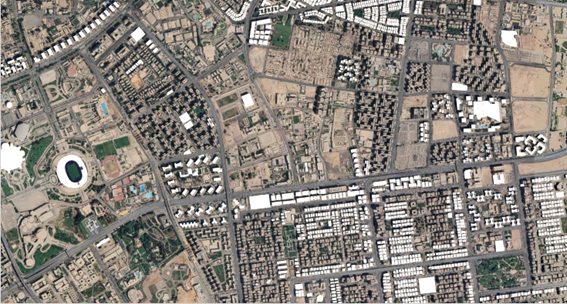}} &
        \frame{\includegraphics[height=.125\linewidth]{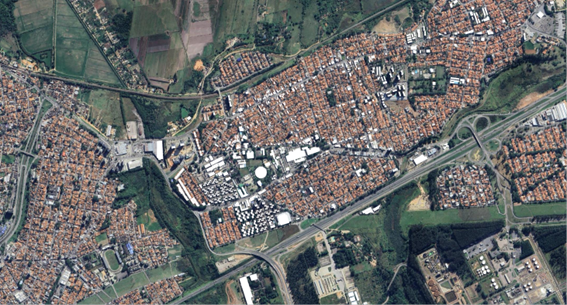}} &
        \frame{\includegraphics[height=.125\linewidth]{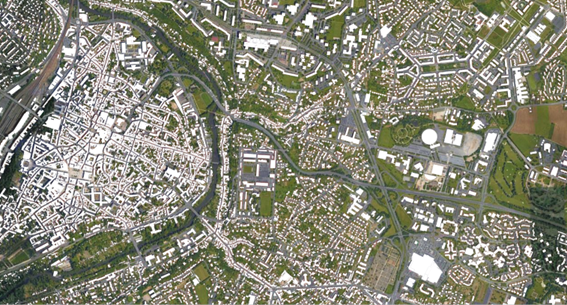}} &
        \frame{\includegraphics[height=.125\linewidth]{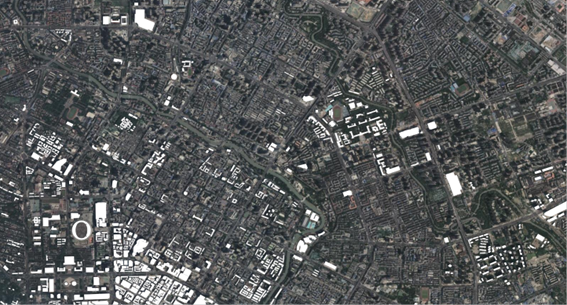}} \\
        \rotatebox[origin=l,y=2mm]{90}{Google} & 
        \frame{\includegraphics[height=.125\linewidth]{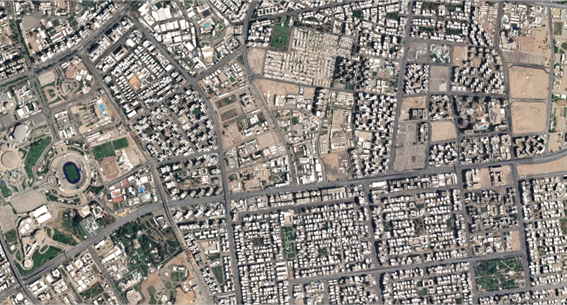}} &
        \frame{\includegraphics[height=.125\linewidth]{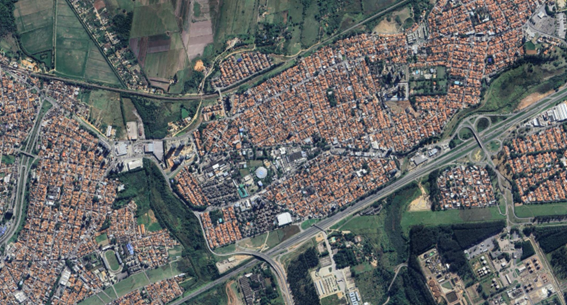}} &
        \frame{\includegraphics[height=.125\linewidth]{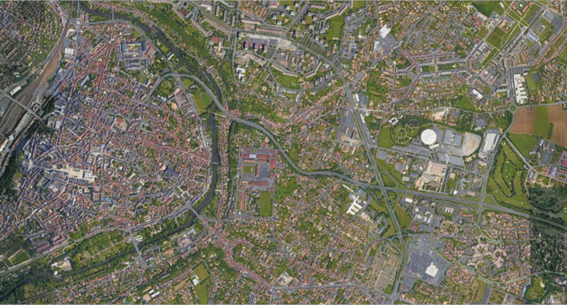}} &
        \frame{\includegraphics[height=.125\linewidth]{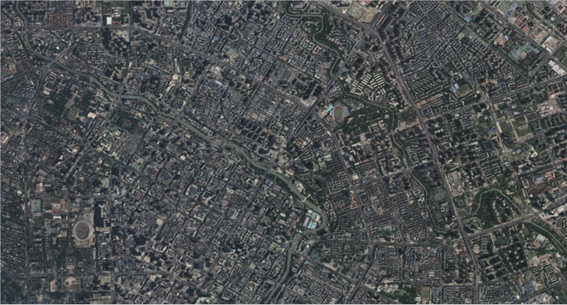}} \\
        \rotatebox[origin=l,y=1mm]{90}{Microsoft} & 
        \frame{\includegraphics[height=.125\linewidth]{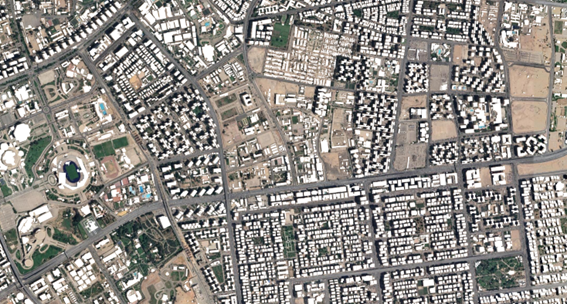}} &
        \frame{\includegraphics[height=.125\linewidth]{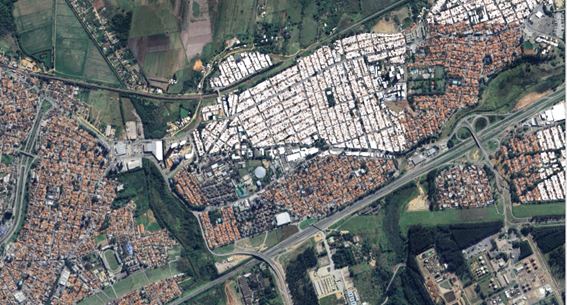}} &
        \frame{\includegraphics[height=.125\linewidth]{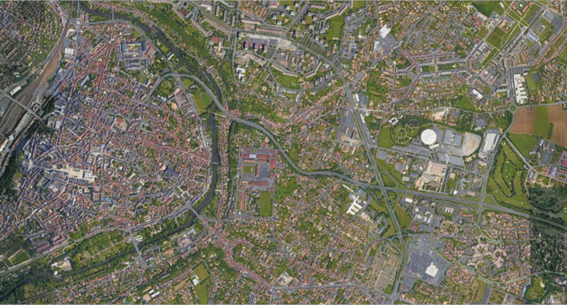}} &
        \frame{\includegraphics[height=.125\linewidth]{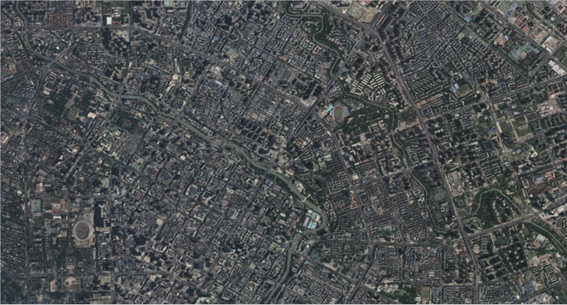}} \\ 
        \rotatebox[origin=l,y=3mm]{90}{GHSL} & 
        \frame{\includegraphics[height=.125\linewidth]{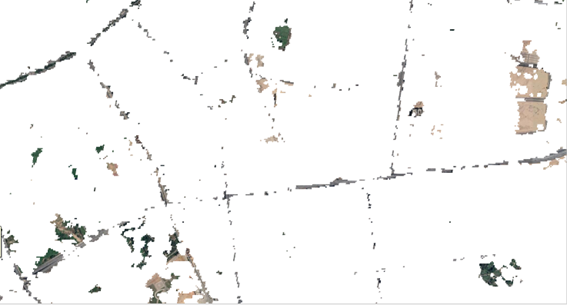}} &
        \frame{\includegraphics[height=.125\linewidth]{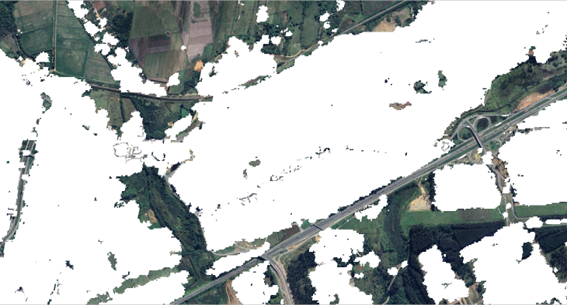}} &
        \frame{\includegraphics[height=.125\linewidth]{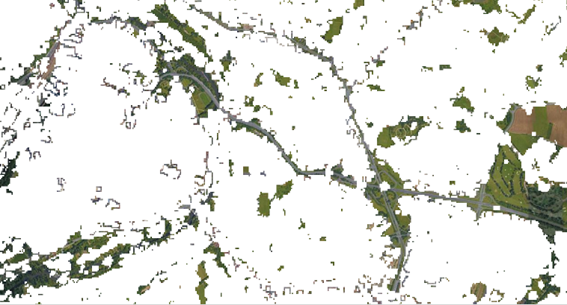}} &
        \frame{\includegraphics[height=.125\linewidth]{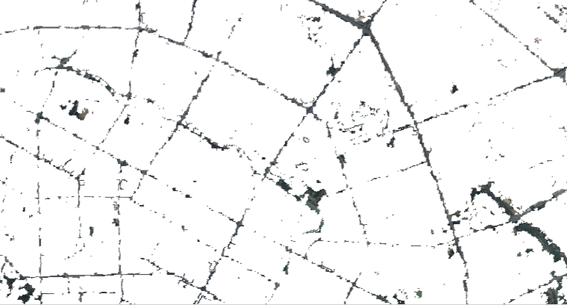}} \\ 
        \rotatebox[origin=l,y=4mm]{90}{WSF} & 
        \frame{\includegraphics[height=.125\linewidth]{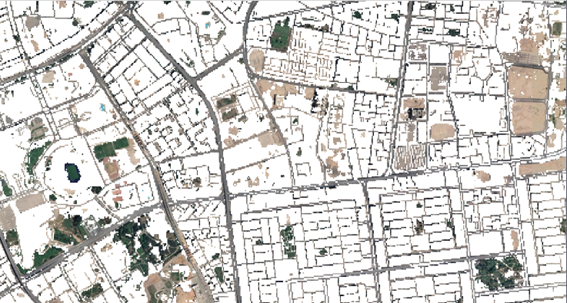}} &
        \frame{\includegraphics[height=.125\linewidth]{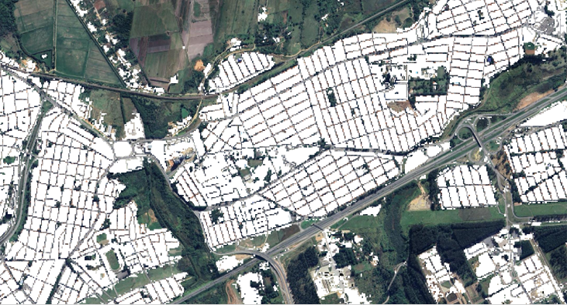}} &
        \frame{\includegraphics[height=.125\linewidth]{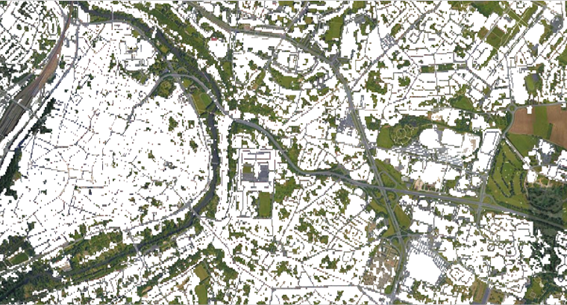}} &
        \frame{\includegraphics[height=.125\linewidth]{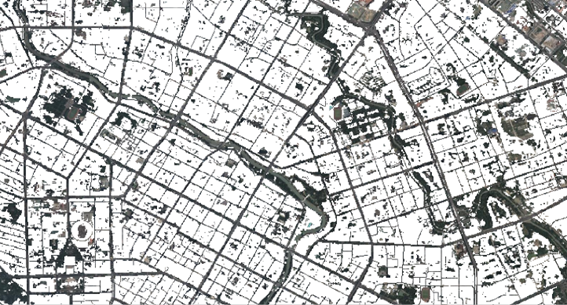}} \\ 
        \rotatebox[origin=l,y=3.5mm]{90}{GBM} & 
        \frame{\includegraphics[height=.125\linewidth]{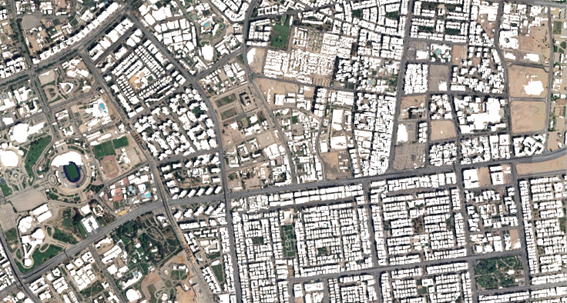}} &
        \frame{\includegraphics[height=.125\linewidth]{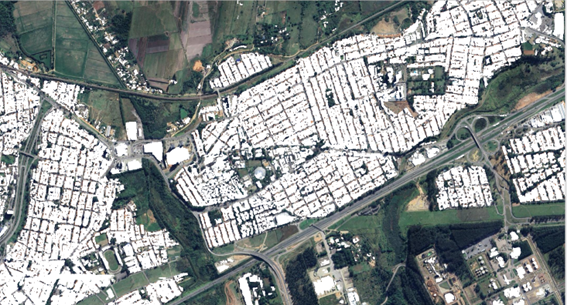}} &
        \frame{\includegraphics[height=.125\linewidth]{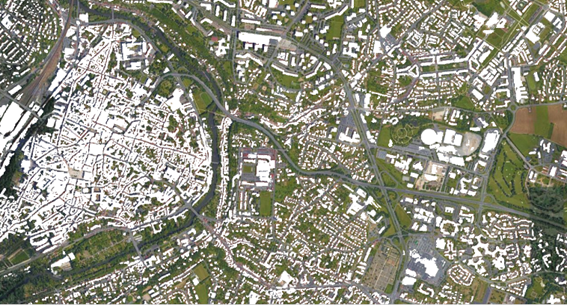}} &
        \frame{\includegraphics[height=.125\linewidth]{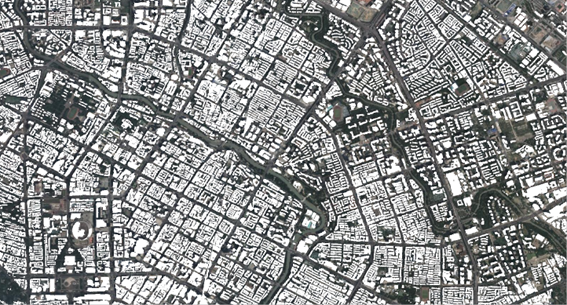}} \\ 
    \end{tabular}
    \caption{Among various data sources, our GBM shows globally consistent building maps on 
a building instance level. From the top to bottom are results (in white) obtained by OSM, Google, Microsoft, GHSL, WSF, and GBM.}
    \label{fig:close}
\end{figure*}

For a closer look at the accuracy of GBM compared to other building products, Fig.~\ref{fig:close} provides a direct comparison for cities from four different continents. OSM has close to complete coverage of Poitiers, France, but only partial coverage of all other cities. Google and Microsoft have good coverage over Cairo, Egypt, but little to no coverage in other regions. Although GHSL and WSF have complete global coverage, they tend to overestimate building area. Only GBM is able to capture the fine details of smaller buildings interspersed by roads and vegetation.

The resulting high quality global building map unveils for the first time the mystery of global buildings, including their global distribution and density. It offers valuable new capabilities in downstream analysis, such as urban structure, solar potential of building areas, vulnerability analysis, population density aggregation, risk assessment, and urban planning. Below, we discuss two specific applications to highlight the value of this dataset.

\section*{Solar Potential Analysis}

The European Green Deal~\cite{fetting2020european} was announced by the European Commission in 2019, which is charged with addressing climate change and promoting sustainable development, with the goal of making the European Union climate neutral by 2050. To this end, fossil fuel and nuclear power plants need to be completely replaced by clean energy. In particular, the recent sanctions related to the conflict between Russia and Ukraine have led to a further energy crisis, confirming the importance of developing new and sustainable energy sources. Also, a recent study by \citet{luderer2022} reported that renewables-based electrification plays a greater role in decarbonization than expected.

Solar energy is clean and renewable. IPCC AR6 shows that solar energy leads all mitigation strategies studied in the report in the potential and cost efficiency of reducing net emission (by 2030) by a significant margin~\cite{ipcc_summary_2022}. Estimates show that on average, 4.5~GtCO\textsubscript{2}-eq per year can be achieved with manageable cost. If every house in the world had solar panels on its roof, would the solar energy generated be sufficient to meet the global energy demand? The question can be easily answered by a joint analysis of the building footprint in GBM and solar potential provided by the Global Solar Atlas. The annual solar potential \(P\) in a given location (\(x,y\)) can be formulated as~\cite{solargis}:
\begin{equation}
    P_{x,y} = \text{PV}_{x,y} * (1 - \text{loss}) * N_p * N_d,
    \label{eq:pv_potential}
\end{equation}
where \(\text{PV}_{x,y}\) is the photovoltaic (PV) power potential of a 1~kW-peak PV system at this location, which is a long-term yearly average of daily totals. \(N_d\) is the number of days in a year, i.e., 365. \(N_p\) is the effective number of 1~kW-peak PV systems, which is defined as \(N_p = A_{b} / A_p\). The term \(A_{b}\) is the total area of global building roofs, whereas \(A_p\) is the area occupied by a 1 kW-peak PV system, which is set in the range of 10--30~m\textsuperscript{2}, depending on the specifications of the solar device. Here we have chosen a range with a reasonable price-performance ratio that would be suitable for large-scale installation. In addition, losses due to dirt and soiling, mismatch, transformer, DC and AC cabling losses, and downtime are considered in the calculation. The average loss for different installation types is set to be 10\%. For more details on our calculations, we recommend interested readers consult \citet{solargis}.

\begin{figure}
    \centering
    \includegraphics[width=\textwidth]{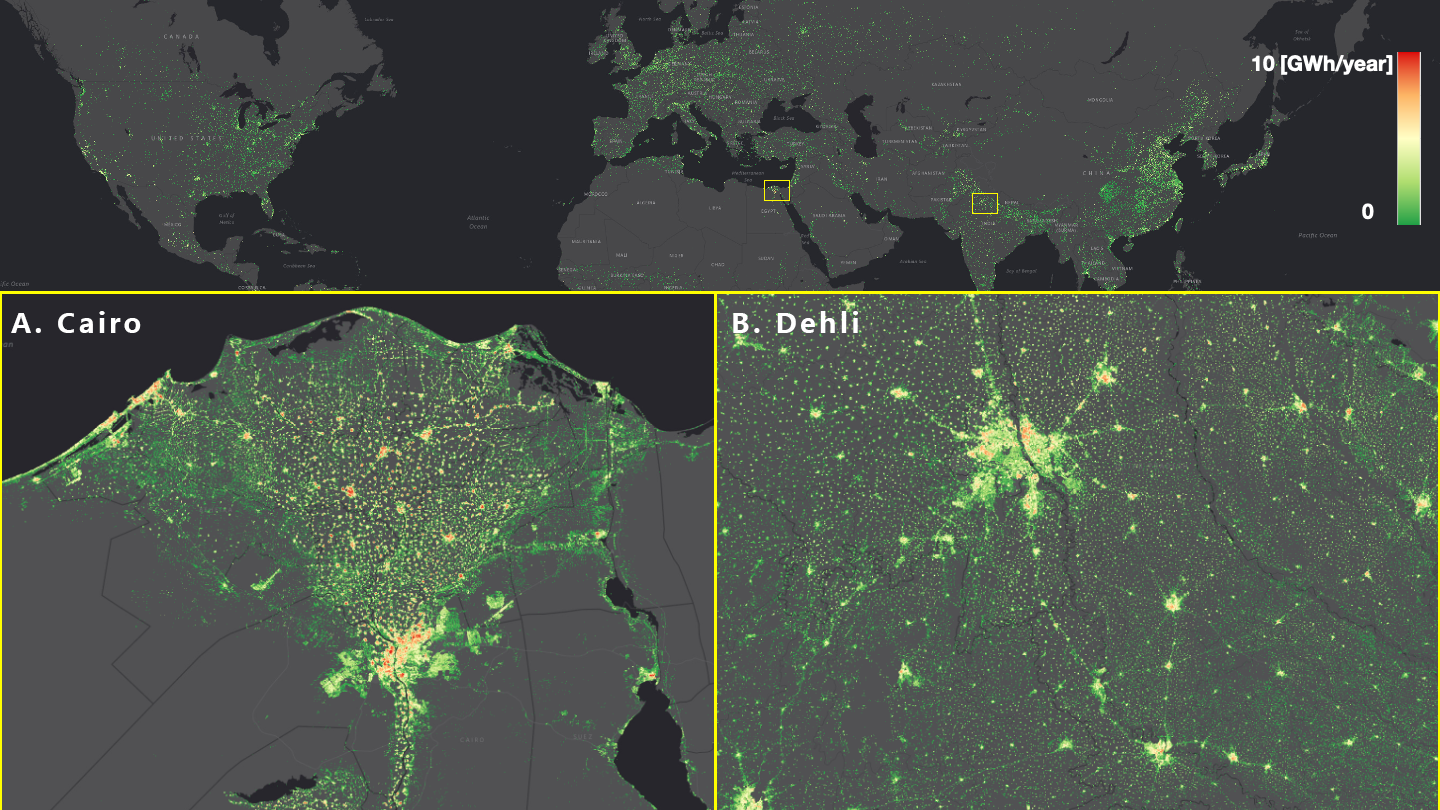}
    \caption{Rooftop solar potential analysis of global buildings. Color indicates the yearly solar potential per pixel in the range of 0--10~GWh/year with a spatial resolution of 250~m~\(\times\)~250~m. Two cities that are ideal for placing rooftop solar panels, A. Cairo, Egypt and B. Delhi, India, are zoomed in.}
    \label{fig:solar}
\end{figure}

Accordingly, we estimate the yearly global rooftop solar potential to be 28--84~PWh, which is 1.1--3.3  times  the  global  energy  consumption  of  2020 --- the  year with the highest energy consumption in human history. Given that the efficiency of the solar panel is assumed to be moderate in this estimate, it would be safe to state that if solar panels were placed on all building roofs, it would be possible to supply the global energy consumption need at a reasonable cost. This offers a promising perspective on promoting rooftop solar PVs across the globe, which currently account for 40\% of the installed capacity of global solar PVs and 25\% of the total renewable capacity additions in 2018~\cite{joshi2021}. Fig.~\ref{fig:solar} shows the resulting global rooftop solar potential map, which reports the yearly solar potential per pixel in the range of 0--10~GWh/year at a spatial resolution of 250~m~\(\times\)~250~m. Two selected areas, Cairo and Delhi, are zoomed in for a better view. Based on such a global map of this type, we can plan and place rooftop solar infrastructures to their best advantage at any place on Earth. Of course, this analysis could only serve as a proof of concept from a macro perspective, for detailed analysis, it is recommended to take into account the orientation of the roof segments and roof superstructures~\cite{li2023solarnet}.

\section*{Socioeconomic and Environmental Variables}

The GBM characterizes the planar dimension of the built structures where people live. Considering that socioeconomic and environmental factors exert a strong influence on the distribution of humans, we aim to investigate whether these factors also have a strong relation to built structures. In this regard, the building area that involves the quantitative information in the planar dimension of buildings is selected as one indicator to further explore the impacts of socioeconomic and environmental variables. Specifically, we choose six variables for regression analysis: population, carbon dioxide (CO\textsubscript{2}) emission, electricity consumption, energy consumption, gross domestic product (GDP), and waste provided by World Bank Open Data~\cite{wb2021}. We examine how those variables relate to the building areas of each country.

\begin{figure}[htbp]
    \centering
    \includegraphics[width=\textwidth]{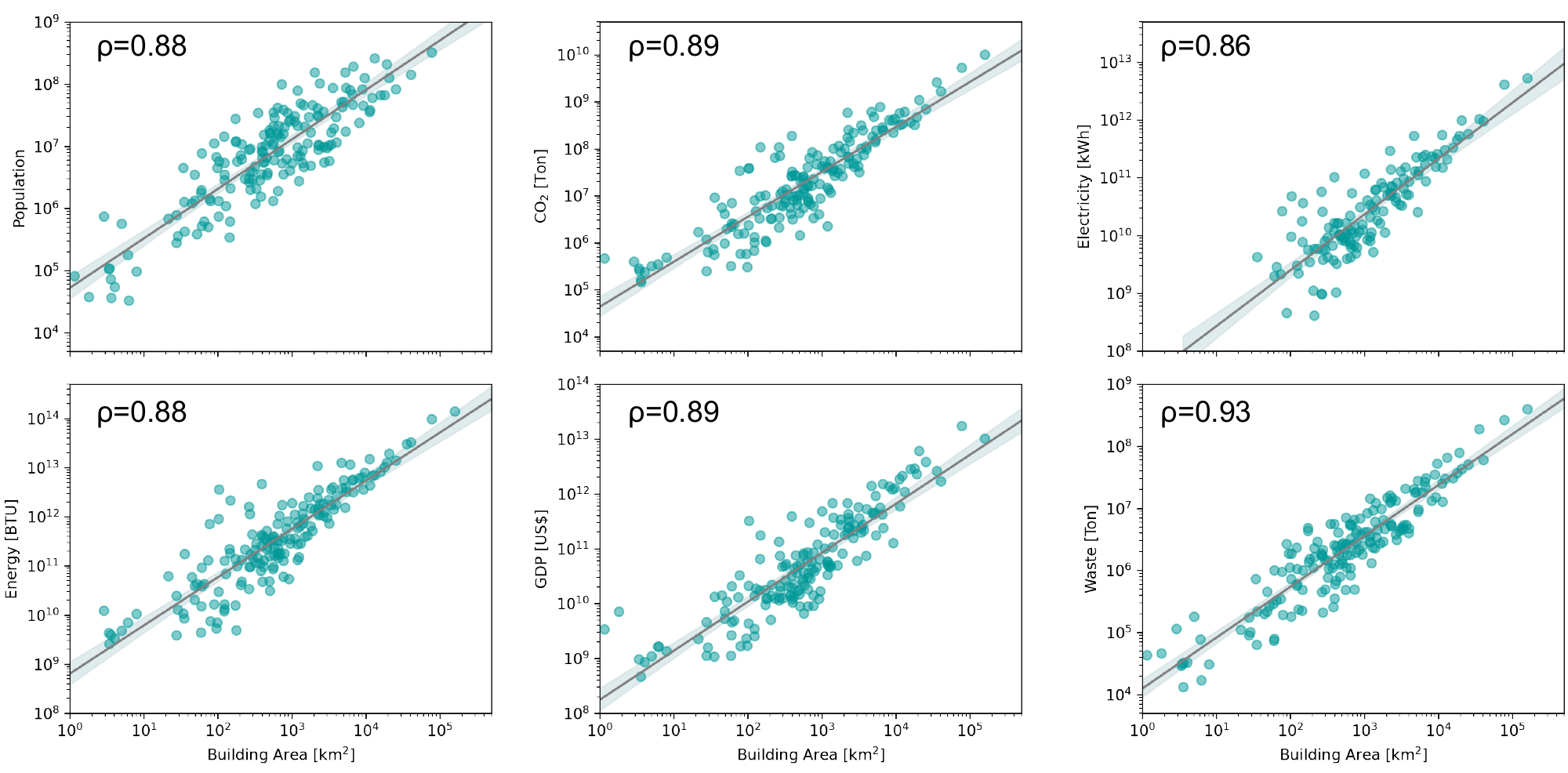}
    \caption{Correlation of building area with six socioeconomic and environmental variables: population, carbon dioxide (CO$_2$) emission, electricity consumption, energy consumption, gross domestic product (GDP), and waste. Each dot represents the total building area and comparison variable for a single country. The correlation coefficient \(\rho\) for each variable is also given.}
    \label{fig:corr}
\end{figure}

In each subplot of Fig.~\ref{fig:corr}, the \(x\)-axis and \(y\)-axis represent the building areas and the socioeconomic and environmental variables of individual countries, respectively. As shown in Fig.~\ref{fig:corr}, building area has a very high positive correlation with all six variables, ranging from 0.86 (electricity) to 0.93 (waste). This confirms that GBM offers a meaningful estimate of characteristics of buildings in relation to socioeconomic and environmental aspects, which can thus serve as an important input to model global socioeconomic needs and drivers. Most existing studies analyze the relationship between building patterns and ecological, environmental, and socioeconomic processes only at the district or city scale due to data limitations~\cite{liu2017landscape,huang2019investigating,lu2021multi}. In contrast, GBM is capable of supporting the research at different scales, e.g., global, continental, national, and city scales.

\section*{Perspectives}

A global map of building footprints can facilitate ongoing efforts to understand human activities, the global urbanization process, socioeconomic phenomena, and the anthropogenic impacts and human risks of climate change. As an example, almost every population density estimation study from national to global scales relies on the data published by the United Nations~\cite{un2019habitat}, the European Union's JRC~\cite{jrc2013jrc}, or Oak Ridge National Laboratory's LandScan~\cite{bhaduri2002landscan}. However, even the United Nations and the national censuses admit that current surveying methods have intrinsic issues, calling the reliability of these studies into question. To date, applied population surveying methods do not incorporate the spatial detail of individual buildings. The best approximation methods so far are random sampling methods and extrapolation models. While current work only qualitatively describes the challenges or shows deviations from official numbers in local sample areas, new building footprints can overcome many shortcomings of these methods. Incorporating such a map of individual buildings can reduce the uncertainty of current population estimates on a global scale.

A further example is analysis of vulnerability to natural hazards and extreme events. Natural hazards, such as floods, earthquakes, landslides, hurricanes, and tsunamis, can cause catastrophic damage and significant socioeconomic loss. For example, record-shattering rainfall caused deadly flooding across Germany and Belgium in July 2021~\cite{thieken2023performance}. Severe floods of similar scale also now frequently take place worldwide, such as in Zhengzhou, China in July 2021~\cite{zhao2023managing}, and in eastern Australia in March 2022~\cite{fryirs2023natural}. Under continued global warming, extreme events will continue to rise in frequency, intensity, duration, and spatial extent over the next decades~\cite{lange2020}. For example, \citet{Thiery2021} estimate that children born in 2020 will experience a two- to sevenfold increase in extreme events, particularly heat waves, compared with people born in 1960, under current climate policy pledges. To accelerate the response to those events, especially for areas with severe damage, it is thus very important to identify vulnerable building areas well in advance. This can be realized by combining geospatial vulnerability data, digital elevation models, and building footprints. Via a few pilot studies, we wish to stimulate wide usage of such unprecedented geoinformation.

Limited to the spatial resolution of global yet affordable satellite data, the accuracy of GBM has room for further improvement, e.g., by employing image super-resolution techniques prior to the segmentation of buildings, especially in places like Africa. In addition, building instances could be segmented along with the building map, which would allow for the polygonization of the footprints of individual buildings. Furthermore, high-resolution building maps updated every five to ten years would be highly relevant for many applications, which would require engagement of a larger science community. Our code and data would offer a good starting point.

Our study demonstrates the high potential and benefits of combining big data acquired by Earth observation (EO) satellites and advanced analytical methods such as deep learning. Going beyond a single use case, this is true for many geographical applications. We live in a golden era of EO with hundreds of petabytes of EO data openly and freely accessible to everyone. This big EO data offers valuable new capabilities in monitoring the changing planet, making predictions with unprecedented spatial and temporal resolution, and providing unique insights into sub-grid-scale processes in the Earth system models that have to be parameterized~\cite{zhu2017deep}. This opportunity, however, comes with serious challenges, related to data analytics, computational cost, and data volumes~\cite{camps2021deep}, to name a few. Further community efforts are needed to develop tailored machine learning/deep learning methods and big data analytic pipelines that consider domain-specific challenges, such as domain shifts across sensors or geographical regions, sparse, imbalanced, and erroneous labels, multisensory data, generalizability and transferability issues, uncertainty quantification, physics-aware machine learning, and computationally-efficient inference. Advances in these key areas will support the full exploitation of the EO data revolution going forward.

\putbib

\end{bibunit}

\newbibstartnumber{39}

\begin{bibunit}

\section*{Methods}

This study establishes a framework for the mapping of global buildings from high-resolution satellite images. This was only possible recently, owing to advances in big data analytics and the availability of high-resolution satellite imagery at relatively low cost. Four convolutional neural networks (CNNs) were utilized to detect buildings in PlanetScope satellite images at 3~m spatial resolution. We collect image patches and publicly-available OpenStreetMap (OSM)~\cite{osm2022planet} building masks from 74 cities across different continents as training data. Due to image cloud coverage and OSM incompleteness or building construction, all patches are manually inspected. Finally, 116,312 quality pairs of satellite image patch and reference building mask are used for the supervised training of four CNN models, which are subsequently used to predict buildings across the entire world. The final GlobalBuildingMap (GBM) is derived from the majority voting of the results predicted by four different models. Additionally, we utilize the resulting GBM for further analysis and derive several raster maps from it. We estimate the global PV power potential that can be derived from global buildings and solar irradiance. We also verify the correlation between building areas and other variables, i.e., socioeconomic and environmental information.

\subsection*{Satellite Data}

We collect nearly 800,000 satellite images from the Planet satellite constellation, which delivers high-resolution (3~m/pixel) optical imagery (red, green, blue, and near-infrared bands) with a high temporal revisiting frequency (up to daily).

\subsubsection*{Urban Detection}

Collection of data over global human settlements is guided by the Global Urban Footprint (GUF)\cite{esch2017breaking}, a global binary map of built-up and non-built-up areas at a 12~m spatial resolution. PlanetScope images are acquired on the basis of a 0.2\degree{} grid. Images are downloaded as long as a single pixel in a 0.2\degree{} grid is human settlement according to GUF. Despite the fact that some settlements might be missing in GUF, we argue that our strategy is able to guarantee a sufficient coverage because of the generous buffer size of 0.2\degree{}.

\subsubsection*{Image Querying}


\begin{table}[htbp]
    \centering
    \begin{tabular}{crr}
        \toprule
        \textbf{Region} & \textbf{Volume [TB]} & \textbf{\# scenes} \\ 
        \midrule
        Africa        & 42  & 211,750\\
        Europe        & 20  & 81,995\\
        North America & 28  & 119,775\\
        South America & 24  & 106,945\\
        Oceania       & 12  & 61,602\\
        East Asia     & 30  & 57,413\\
        West Asia     & 13  & 139,592\\
        \midrule
        World         & 169 & 779,072 \\
        \bottomrule
    \end{tabular}
    \caption{Volume and number of PlanetScope satellite scenes for different continents.}
    \label{tab:data}
\end{table}

Only PlanetScope images with a cloud and haze percentage less than 10\% are used in this study. The majority of queried images are acquired from 2019. When insufficient images of required quality are available in 2019, images from 2018 are downloaded instead. In the rare case where cloud- and haze-free imagery is unavailable for both 2019 and 2018, a lower resolution basemap is used instead. The volume and number of scenes of the acquired data are given in Table~\ref{tab:data}.

\subsection*{Analysis-Ready Data}

The spectral values in each band of PlanetScope imagery come in a wide dynamic range among different cities. Extreme values of pixel intensity and highly compressed image histograms often occur. This is caused by a variety of reasons including different illumination conditions and different building construction materials. Considering that unscaled input variables can result in a slow or unstable learning process~\cite{bishop1995neural}, we implement a radiometric calibration method based on robustly scaling and clipping the histogram of each spectral band. In order to retain as much information about the histogram as possible, we robustly estimate the scale and the main body of the distribution using the interquartile range (IQR). IQR is the difference between the 75th (\(Q_3\)) and 25th (\(Q_1\)) quartiles of the distribution, and is a commonly used robust measure of scale. For example, \(Q_1 - 1.5 \cdot \text{IQR}\) and \(Q_3 + 1.5 \cdot \text{IQR}\) correspond to approximately $\mu \pm 3 \sigma$ in a normal distribution. For heavily-tailed distributions like some bands in the Planet imagery, we will be able to tell the bulk of the pixel values and then clip off the extreme values. Since the distribution of the pixel value is primarily heavily tailed, the clipping range was set from 0 to \(Q_3 + 1.5 \cdot \text{IQR}\) in our radiometric calibration, meaning only extremely large values are clipped. Afterward, all pixel values are normalized to the range 0--1. By clipping extreme values using IQR instead of a fixed percentile, we are able to adaptively control the amount of information being discarded. Our experiments show its robustness for data preprocessing at a large scale where satellite imagery covers the majority of the globe. Finally, we mosaic all images and crop them to individual tiles with size of 5\degree{} in latitude and longitude.

\subsection*{Machine Learning Pipeline}

Below, we detail the deep learning pipeline used to train the model.

\subsubsection*{Training and Validation Data Preparation}

\begin{figure}[htbp]
    \centering
    \includegraphics[width=\textwidth]{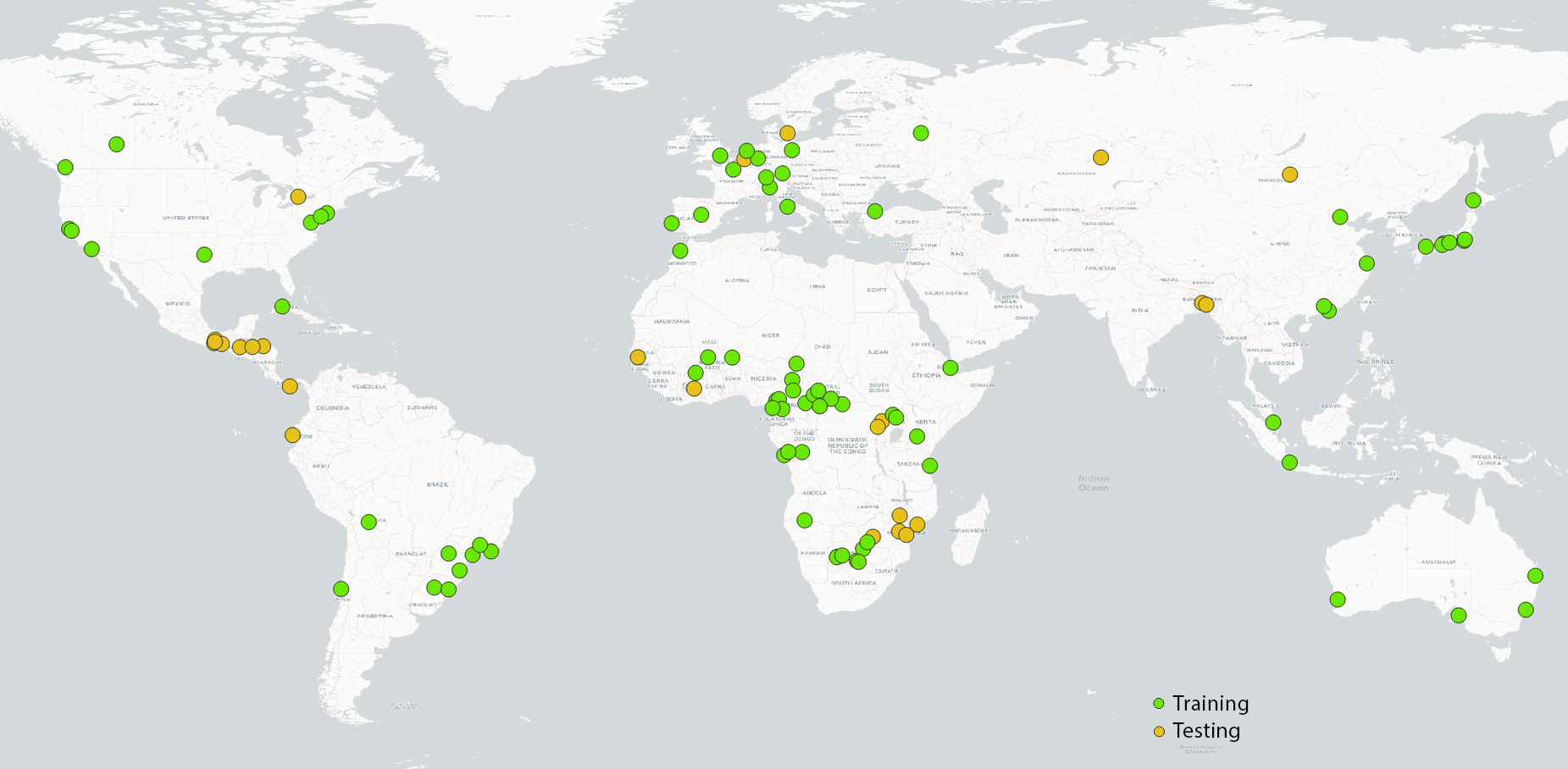}
    \caption{The 74 training and 34 testing cities used in this work. The data are distributed over all the continents except Antarctica.}
    \label{fig:planetscope}
\end{figure}

For training data preparation, we select Planet satellite imagery and the corresponding building footprints (stored as polygon shapefiles) from OpenStreetMap, where the detailed building footprints around 74 cities (see  Fig.~\ref{fig:planetscope}) are publicly released. Vector-format building masks are then rasterized to the same resolution as the corresponding satellite imagery for each city. Together with PlanetScope imagery, the rasterized building footprints are cropped to patches of 256~\(\times\)~256~pixels. 

We observe two major issues of using OSM building footprints and PlanetScope images as training pairs. The first is the mismatch of the two in some areas due to the incorrectness and incompleteness of OSM building footprints. For example, a building might appear in the satellite image, while it is missing in the corresponding OSM building footprints, or vice versa. Therefore, those erroneous pairs were discarded via manual inspection. The second issue is the misalignment between certain pairs, as building footprints in OSM are derived from various data sources that are different in geolocation accuracy. Therefore, coregistration was implemented in order to align the satellite imagery with the corresponding building footprints in OSM. To find the translation, satellite imagery is first transformed into a grayscale image, and a Sobel operator~\cite{sobel1990isotropic} is then applied to both the grayscale satellite image and its corresponding building footprints. Finally, a cross-correlation is calculated between the two, and the maximum of the cross-correlation indicates the offsets between these two data sources, which correspond to the translation coefficients. The dataset is then separated into two parts, where 80\% of the sample patches are used for training the network and 20\% (23,287) are used for model validation. 

Instead of the typical binary class label (1 as ``building'' and 0 as ``non-building''), a truncated signed distance label is utilized as the target representation for model training. By doing so, both semantic information and geometric properties of the buildings can be learned~\cite{bischke2017multi}. Specifically, the truncated signed distance label takes the signed distances from pixels to the boundaries of buildings into consideration, where positive values indicate building interior and negative means building exterior. Afterward, the distance is truncated at a given threshold \(\beta\), only incorporating the pixels closest to building boundaries~\cite{bischke2017multi}. Here, \(\beta\) is empirically set as 10. Finally, the distance values are categorized into a number of class labels~\cite{bischke2017multi}. For instance, 11 classes with the labels \(L=\{ 0,1,2,...,10 \}\) are categorized in our research. When the class label is larger than 5, this pixel belongs to the building, and vice-versa. The truncated signed distance label is able to capture both semantic information and implicit geometric properties of each pixel.

\subsubsection*{Convolutional Neural Networks}

In our research, the task of building footprint generation belongs to the branch of semantic segmentation in computer vision, which assigns each pixel with a label (building or non-building). Convolutional Neural Networks (CNNs) effectively solve semantic segmentation problems, as they are capable of learning an enhanced feature representation directly from raw input. CNNs can get rid of heuristic feature design procedures and achieve better generalization capabilities than other traditional methods. Encoder-decoder-based networks are commonly used architectures, where satellite imagery is mapped into highly efficient feature representations in the encoder and then recovered into a segmentation map in the decoder. In our research, three encoder-decoder-based networks are selected: Graph CNN~\cite{shi2020building}, FC-DenseNet~\cite{jegou2017one}, and Eff-UNet~\cite{baheti2020eff}, which have proven to be successful for the task of building footprint generation. Graph CNN integrates the Graph Convolutional Network (GCN) and Deep Structured Feature Embedding (DSFE) into an end-to-end workflow, where DFSE is utilized to extract more representative features and GCN aggregates the information from neighbor pixels to learn about local structures. More specifically, FC-DenseNet as a feature extractor in DSFE provides comprehensive features for a gated GCN that is proposed to model both local and global contextual dependencies. This helps to preserve sharp building boundaries and fine-grained pixel-level predictions. Both the encoder and decoder in FC-DenseNet are composed of five dense blocks, and each dense block has five convolutional layers. The key element of FC-DenseNet is the DenseNet~\cite{huang2017densely} block, which combines features using iterative concatenation. By doing so, a more efficient flow of information can be provided through network learning to improve semantic segmentation results. The encoder of Eff-UNet is EfficientNet~\cite{tan2019efficientnet}, which can efficiently learn feature maps. The decoder of Eff-UNet is comprised of five transposed convolutional layers that upsample the convolved image to predict segmentation masks. The advantages of Eff-UNet can be attributed to its capability of systematically improving performance with all compound coefficients of the architecture (width, depth, and image resolution) balanced. Moreover, it consumes much less training and inference time than the other two competitors.

\subsubsection*{Network Training}

The above-mentioned three networks are trained using \textit{per city}-normalized training data. The FC-DenseNet network was additionally trained using \textit{per patch}-normalized training data (i.e., the radiometric calibration was applied on individual patches). This approach may be beneficial to the inference stage where the spectral range of individual cities is unknown. In total, four CNN models are trained. All models were trained for 150 epochs, using a stochastic gradient descent (SGD) optimizer with a learning rate of \(10^{-5}\). The training batch size of all models is set as 4. Negative log-likelihood loss was used as the loss function for all models.

\subsection*{Inferencing}

Below, we detail the inference pipeline used to create the final GBM predictions.

\subsubsection*{Parallel processing}

Since the volume of data to be processed is extremely large, a distributed data-parallelization (DDP) scheme is applied during inference time. We first compute the cross product between four different models and all input images. Each combination is then used to form process pools, and each process is assigned to a GPU in round robin fashion. In order to further speed up the processing, the batch size during inferencing is maximized according to the specification of the GPU accelerator. Meanwhile, the number of workers is optimized by PyTorch. Eight NVIDIA GPUs were employed for inferencing.

\subsubsection*{Model Ensemble}

Instead of relying on a single model to make predictions, we use an ensemble of all four models. Final determination of predicted class is made by the majority vote between all models. A pixel is classified as building if and only if two or models predict it as such. By doing so, we mitigate the randomness of model training, as well as enhance the generalization capability of the models. This is especially beneficial for areas with high prediction uncertainty.

\subsection*{Post-Processing}

In order to reduce the misclassified building pixels, two land cover layers were applied to the predicted global building maps. One is the WSF layer, which is the binary land cover mask of urban and non-urban. The WSF layer is used to coarsely distinguish the urban and non-urban areas. Further, another layer, namely ``finer resolution observation and monitoring-global land cover'' (FROM-GLC10)~\cite{chen2019stable} is applied for filtering, which contains nine classes, including cropland, forest, grass, shrub, water, impervious, bare land, snow, and cloud. Since false positives mainly exists in non-urban areas, we introduce area-aware filtering criteria. For urban areas, a weak filter is applied, which only filters out classes like cropland, grass, and shrub. For non-urban areas, all classes are used for filtering except for the impervious class.

\subsection*{Evaluation}

Below, we describe the accuracy of GBM predictions using both quantitative and qualitative metrics.

\subsubsection*{Statistical Comparison}


The performance of the models and their ensemble are evaluated by two metrics, F1 score and intersection over union (IoU), over the validation set in 74 cities. Although FC-DenseNet with per-patch normalization has the weakest performance on the validation set, it actually has the strongest performance on the test set. We theorize that this is due to large brightness variations within cities. The train and test sets are sourced from the same 74 cities and divided by random split, so the model has seem images from the same cities before. However, the test cities represent a different geographic split, and are more useful in gauging model performance in unseen regions. We found the FC-DenseNet model to be superior to GCN-FSFE and Eff-UNet in our experiments.

\begin{sidewaystable}[htbp]
    \centering
    \caption{Accuracy metrics of different building map products evaluated against OpenStreetMaps data across cities distributed over five continents. Note that Microsoft and Google do not offer global coverage, resulting in lower overall accuracy.}
    \begin{tabular}{ccrcccccccccc}
        \toprule
        & & & \multicolumn{2}{c}{\textbf{GBM}} & \multicolumn{2}{c}{\textbf{Microsoft}} & \multicolumn{2}{c}{\textbf{Google}} &  \multicolumn{2}{c}{\textbf{WSF}} & \multicolumn{2}{c}{\textbf{GHSL}} \\
        \textbf{Continent} & \textbf{City} & \textbf{\# patches} & \textbf{F1 score} & \textbf{IoU} & \textbf{F1 score} & \textbf{IoU} & \textbf{F1 score} & \textbf{IoU} & \textbf{F1 score} & \textbf{IoU} & \textbf{F1 score} & \textbf{IoU} \\
        \midrule
        \multirow{11}{*}{Africa} 
          & Bangui & 307 & 0.35 & 0.21 & 0.42 & 0.27 & 0.49 & 0.33 & 0.36 & 0.22 & 0.29 & 0.17 \\
        ~ & Beira & 427 & 0.33 & 0.20 & 0.63 & 0.46 & 0.45 & 0.29 & 0.34 & 0.20 & 0.25 & 0.14 \\
        ~ & Beni & 86 & 0.30 & 0.18 & 0.66 & 0.50 & 0.53 & 0.36 & 0.34 & 0.20 & 0.27 & 0.16 \\
        ~ & Bouake & 374 & 0.41 & 0.26 & 0.58 & 0.41 & 0.60 & 0.43 & 0.41 & 0.26 & 0.31 & 0.19 \\
        ~ & Bulawayo & 1,151 & 0.35 & 0.21 & 0.41 & 0.26 & 0.37 & 0.22 & 0.37 & 0.23 & 0.21 & 0.12 \\
        ~ & Bunia & 54 & 0.32 & 0.19 & 0.41 & 0.26 & 0.50 & 0.33 & 0.32 & 0.19 & 0.26 & 0.15 \\
        ~ & Chimoio & 114 & 0.27 & 0.15 & 0.43 & 0.27 & 0.38 & 0.23 & 0.29 & 0.17 & 0.22 & 0.12 \\
        ~ & Cobán & 123 & 0.48 & 0.31 & - & - & 0.48 & 0.32 & 0.51 & 0.34 & 0.21 & 0.12 \\
        ~ & Farafenni & 22 & 0.51 & 0.34 & 0.63 & 0.46 & 0.55 & 0.38 & 0.36 & 0.22 & 0.27 & 0.16 \\
        ~ & Quelimane & 62 & 0.44 & 0.28 & 0.30 & 0.18 & 0.48 & 0.31 & 0.33 & 0.20 & 0.27 & 0.15 \\
        ~ & Tete & 111 & 0.29 & 0.17 & 0.45 & 0.29 & 0.49 & 0.32 & 0.29 & 0.17 & 0.23 & 0.13 \\
        ~ & Overall & 2,831 & 0.36 & 0.22 & 0.44 & 0.28 & 0.47 & 0.31 & 0.36 & 0.22 & 0.25 & 0.14 \\
        \midrule
        \multirow{4}{*}{Asia}
          & Astana & 654 & 0.52 & 0.36 & 0.50 & 0.33 & - & - & 0.39 & 0.24 & 0.19 & 0.11 \\
        ~ & Beijing & 456 & 0.58 & 0.41 & - & - & - & - & 0.47 & 0.31 & 0.35 & 0.21 \\
        ~ & Cumilla & 112 & 0.45 & 0.29 & 0.50 & 0.34 & 0.45 & 0.29 & 0.50 & 0.34 & 0.38 & 0.23 \\
        ~ & Dhaka & 960 & 0.54 & 0.37 & 0.47 & 0.31 & 0.49 & 0.33 & 0.57 & 0.39 & 0.45 & 0.29 \\
        ~ & Guangzhou & 500 & 0.61 & 0.44 & - & - & - & - & 0.58 & 0.41 & 0.47 & 0.31 \\
        ~ & Shanghai & 500 & 0.57 & 0.39 & - & - & - & - & 0.47 & 0.31 & 0.36 & 0.22 \\
        ~ & Ulaanbaatar & 87 & 0.46 & 0.30 & 0.59 & 0.41 & - & - & 0.43 & 0.27 & 0.35 & 0.21 \\
        ~ & Overall & 3,269 & 0.55 & 0.38 & 0.33 & 0.20 & 0.28 & 0.16 & 0.51 & 0.34 & 0.35 & 0.21 \\
        \midrule
        \multirow{2}{*}{Europe} 
          & Brussels & 1,808 & 0.64 & 0.47 & 0.44 & 0.28 & - & - & 0.60 & 0.43 & 0.39 & 0.24 \\
        ~ & Copenhagen & 797 & 0.64 & 0.47 & 0.71 & 0.55 & - & - & 0.54 & 0.37 & 0.36 & 0.22 \\
        ~ & Overall & 2,605 & 0.64 & 0.47 & 0.55 & 0.38 & - & - & 0.58 & 0.41 & 0.38 & 0.23 \\
        \midrule
        North America 
          & Dallas & 500 & 0.61 & 0.44 & 0.79 & 0.66 & - & - & 0.47 & 0.31 & 0.25 & 0.14 \\
        ~ & Edmonton & 500 & 0.65 & 0.48 & 0.88 & 0.78 & - & - & 0.50 & 0.33 & 0.35 & 0.21 \\
        ~ & Philadelphia & 500 & 0.69 & 0.53 & 0.83 & 0.70 & - & - & 0.60 & 0.43 & 0.43 & 0.27 \\
        ~ & San Jose & 500 & 0.62 & 0.44 & 0.92 & 0.84 & - & - & 0.53 & 0.36 & 0.38 & 0.23 \\
        ~ & Toronto & 500 & 0.60 & 0.43 & 0.78 & 0.64 & - & - & 0.52 & 0.35 & 0.34 & 0.21 \\
        ~ & Vancouver & 500 & 0.57 & 0.40 & 0.85 & 0.74 & - & - & 0.43 & 0.27 & 0.28 & 0.16 \\
        ~ & Overall & 3,000 & 0.63 & 0.46 & 0.85 & 0.73 & - & - & 0.52 & 0.35 & 0.34 & 0.20 \\
        \midrule
        \multirow{8}{*}{South America}
          & Juchitán & 140 & 0.48 & 0.31 & - & - & 0.58 & 0.40 & 0.45 & 0.29 & 0.35 & 0.21 \\
        ~ & Matias Romero Avendano & 43 & 0.51 & 0.34 & - & - & 0.43 & 0.28 & 0.41 & 0.26 & 0.27 & 0.16 \\          
        ~ & Portoviejo & 117 & 0.47 & 0.31 & 0.56 & 0.39 & 0.56 & 0.39 & 0.48 & 0.32 & 0.35 & 0.21 \\
        ~ & Salina Cruz & 238 & 0.43 & 0.28 & - & - & 0.57 & 0.40 & 0.43 & 0.27 & 0.29 & 0.17 \\
        ~ & San Pedro Sula & 730 & 0.54 & 0.37 & - & - & 0.61 & 0.44 & 0.50 & 0.33 & 0.39 & 0.24 \\
        ~ & Santiago de Veraguas & 61 & 0.52 & 0.35 & 0.70 & 0.54 & 0.75 & 0.60 & 0.55 & 0.38 & 0.39 & 0.24 \\
        ~ & Tocoa & 39 & 0.48 & 0.32 & - & - & 0.59 & 0.42 & 0.44 & 0.28 & 0.33 & 0.20 \\
        ~ & Tonalá & 45 & 0.41 & 0.26 & - & - & 0.49 & 0.33 & 0.47 & 0.31 & 0.35 & 0.21 \\
        ~ & Overall & 1,413 & 0.50 & 0.33 & 0.14 & 0.08 & 0.59 & 0.42 & 0.48 & 0.31 & 0.36 & 0.22 \\
        \midrule
        World & Overall & 13,118 & 0.55 & 0.38 & 0.53 & 0.36 & 0.28 & 0.17 & 0.49 & 0.32 & 0.34 & 0.21 \\
        \bottomrule
    \end{tabular}
    \label{tab:experiment}
\end{sidewaystable}

We made further statistical comparison between our results and the building maps available from Microsoft~\cite{microsoft2023building}, Google~\cite{sirko2021continental}, WSF~\cite{marconcini2021understanding}, and GHSL~\cite{pesaresi2023ghs}. The results are shown in Table~\ref{tab:experiment}. Although Microsoft and Google use higher resolution imagery and offer higher accuracies in many cities, neither product offers global coverage. GBM offers a higher resolution and higher overall accuracy than WSF and GHSL for 4 out of 5 continents, and offers equal accuracy in Africa. GBM offers a notable advantage over Microsoft and Google in Europe and Asia, but falls behind in North America where Microsoft has put much of their focus.

\subsubsection*{Visual Comparison}

\begin{figure}[htbp]
    \centering
    \includegraphics[width=\textwidth]{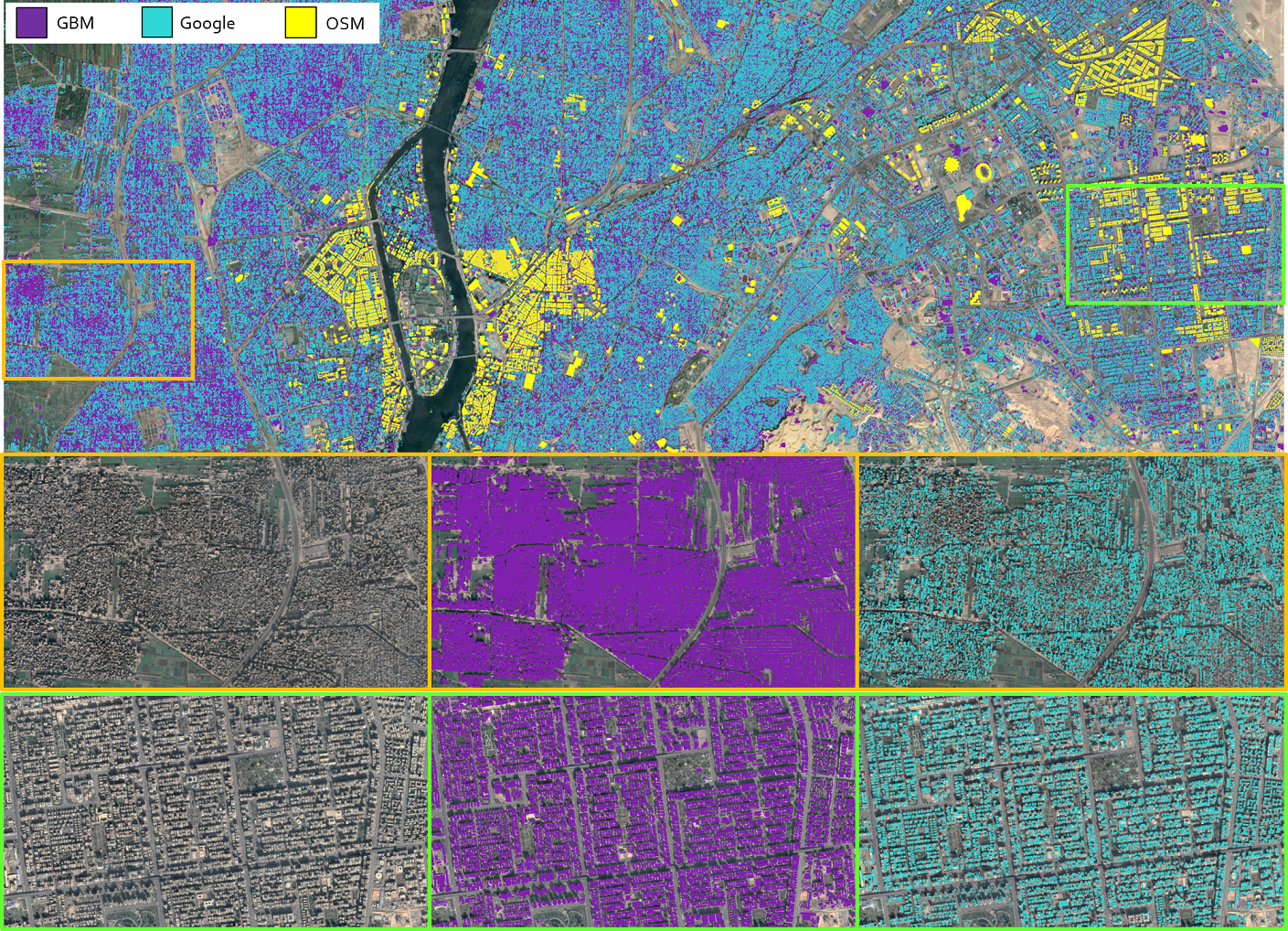}
    \caption{Visual comparison of building footprints from different data sources in Cairo, Egypt. The three building footprint layers from GBM (purple), Google (cyan) and OSM (yellow) are overlaid with high-resolution aerial image. Two selected areas, i.e., dense area/informal settlement (orange) and non-dense area (green) are zoomed in. Each area has three subfigures, which show the corresponding high-resolution aerial image as reference (left), GBM overlaid with satellite image (mid) and Google overlaid with satellite image (right).}
    \label{fig:cairo}
\end{figure}

As a representative city in Africa, Cairo, Egypt is selected to visually exemplify building footprints from GBM, OSM, and Google in Fig.~\ref{fig:cairo}. In order to give a direct visual reference to where the buildings are, we use a higher resolution aerial image as the background image on which we overlay GBM (purple), Google (cyan), and OSM (yellow). Two areas, dense (orange) and non-dense (green), are zoomed in to further detail the quality of the building footprints. As expected, OSM has only limited information and does not have building footprints for most of the buildings. For a more planned area, while comparing to the higher resolution optical image (left), both GBM and Google offer very accurate building layers. Nevertheless, it can be observed that the Google building footprints miss a large number of buildings in the dense area, while GBM properly segments these buildings, despite being trained on lower resolution PlanetScope data.

\subsection*{Solar Potential Calculation}

The yearly solar potential was calculated using World Bank's Solargis raster datasets~\cite{solargis}. These datasets are available for any location between latitudes 60\textdegree{}N and 50\textdegree{}S with a resolution of 1~km. The daily solar potential outside the latitudes was assumed with a constant value of 3.5~kWh/kWp (peak)~\cite{joshi2021}. The potential losses caused by different factors are considered according to the Solargis model. Finally, the value of every building pixel is aggregated to compute the yearly solar potential of global buildings.

\subsection*{Regression Analysis}

To offer a holistic analysis of our GBM, we select a number of socioeconomic and environmental variables to explore their relationships with morphological building parameters (i.e., building area in our study). Specifically, six variables --- population, carbon dioxide (CO\textsubscript{2}) emission, electricity consumption, energy consumption, gross domestic product (GDP), and waste --- are sourced from World Bank Open Data~\cite{wb2021}, in order to carry out regression analysis. 

We first compute the statistics of building area as well as socioeconomic and environmental variables for each country. Afterward, we apply a linear regression model for assessing the relationship between building area and each of the six variables. Experiments show that all variables have a significant positive correlation with building area. This shows GBM can play a significant role in supporting related socioeconomic and environmental process studies. Humans construct buildings for continuous occupancy. Hence, building footprints can effectively reflect the pattern of human settlements. In other words, the statistics of buildings (e.g., building area) derived from building footprint maps can have a close relationship with the population. Larger populations contribute to increased greenhouse gas emissions as well as electricity and energy consumption. The impact of economic level on the spatial distribution of building area has been shown to be significant~\cite{liu2021high}. The rising GDP that results from the concentration of economic activities can accelerate the construction of new buildings.

\bmhead{Data and Materials Availability}

The code, as well as the resulting building footprint map and rooftop solar potential map, will be made open access after acceptance.

\renewcommand\refname{Methods References}

\putbib

\end{bibunit}

\backmatter

\bmhead{Acknowledgments}

The authors would like to thank Richard Bamler and Jonathan Bamber for reviewing the draft. Source of the Planet data used in the publication: Planet Labs Inc. The authors acknowledge Fahong Zhang for his support in data downloading. The work is jointly supported by the European Research Council (ERC) under the European Union's Horizon 2020 research and innovation programme (grant agreement No. [ERC-2016-StG-714087], Acronym: \textit{So2Sat}), by the German Federal Ministry of Education and Research (BMBF) in the framework of the international future AI lab ``AI4EO -- Artificial Intelligence for Earth Observation: Reasoning, Uncertainties, Ethics and Beyond" (grant number: 01DD20001), by the German Federal Ministry for Economic Affairs and Climate Action in the framework of the ``national center of excellence ML4Earth" (grant number: 50EE2201C), by the Excellence Strategy of the Federal Government and the Länder through the TUM Innovation Network EarthCare, and by the Munich Center for Machine Learning. The authors gratefully acknowledge the Gauss Centre for Supercomputing e.V. (\url{www.gauss-centre.eu}) for funding this project by providing computing time through the John von Neumann Institute for Computing (NIC) on the GCS Supercomputer JUWELS at Jülich Supercomputing Centre (JSC) and on SuperMUC-NG at Leibniz Supercomputing Centre (www.lrz.de). 

\bmhead{Authors Contributions}

Conceptualization: X.Z.; methodology: X.Z., Y.S., Q.L., Y.W.; software: Y.S., Q.L., Y.W.; results validation: X.Z., Y.S., Q.L., Y.W.; analysis: X.Z., Y.S., Q.L., Y.W.; data collection: X.Z., Y.S., Q.L., Y.W.; writing, and original draft preparation: X.Z., Q.L., Y.L., Y.W.; paper revision: A.S., X.Z., Q.L., Y.L., Y.W.; visualization: Y.W., X.Z., Q.L., Y.L.; funding acquisition: X.Z.; Project administration: X.Z., Y.W.; Resources: X.Z.. J.P. developed the Planet data downloading script and downloaded the Planet data during his TUM affiliation.

\bmhead{Competing Interests}

The authors declare no competing interests.

\bmhead{Additional Information}

Supplementary Information is available for this paper.

Correspondence and requests for materials should be addressed to X.Z.

Reprints and permissions information is available at \url{www.nature.com/reprints}.

\section*{Supplementary Information}

\subsection*{Definitions}

One of the issues with cross-comparison of different building maps is the lack of clarity on what constitutes a ``building''. Most of the above-mentioned papers either have conflicting definitions of buildings or no definitions at all. These conflicting definitions often pertain to whether or not temporary shelters and slums are included, and are of vital importance to a number of potential applications. Below, we describe what information we could find about the definitions used in each product.

Global Urban Footprint (GUF) is a map of \emph{built-up area}, which they define as ``a region featuring man-made building structures with a vertical component''. High-Resolution Settlement Layer (HRSL) is a map of \emph{settled} regions, which they define as ``containing buildings''. Google and Microsoft offer no definitions of what they consider to be a building. OpenStreetMaps (OSM) explicitly excludes temporary structures, and defines a \emph{building} as ``a man-made structure with a roof, standing more or less permanently in one place''.

World Settlement Footprint (WSF) explicitly acknowledges the above-mentioned issues with unclear definitions. They use the following definitions in their WSF-2015 paper:

\begin{description}
    \item[building:] ``any structure having a roof supported by columns or walls and intended for the shelter, housing, or enclosure of any individual, animal, process, equipment, goods, or materials of any kind''
    \item[building lot:] ``the area contained within an enclosure (e.g., wall, fence, hedge) surrounding a building or a group of buildings''
    \item[road:] ``any long, narrow stretch with a smoothed or paved surface, made for traveling by motor vehicle, carriage, etc., between two or more points''
    \item[paved surface:] ``any level horizontal surface covered with paving material''
\end{description}

WSF further acknowledges that different papers use different definitions of \emph{settlement}, and provides performance metrics for three different possible definitions:

\begin{enumerate}
    \item buildings
    \item buildings + building lots
    \item buildings + building lots + roads/paved surfaces
\end{enumerate}

However, the authors fail to mention which of these three definitions is used in the final product.

Among the related works, Global Human Settlement Layer (GHSL) has the most clear definitions used in their products. They use the following definitions:

\begin{description}
    \item[building] ``any roofed structure erected above ground for any use''
    \item[built-up surface / building footprint] ``the gross surface (including the thickness of the walls) bounded by the building wall perimeter with a spatial generalization matching the 1:10K topographic map specifications''
    \item[built-up fraction] ``the share of the raster sample (i.e. pixel or grid cell) surface that is covered by the built-up surface''
    \item[residential domain / residential use] ``the built-up surface dedicated prevalently for residential use''
    \item[non-residential domain] ``any built-up surface not included in the [residential domain] class''
\end{description}

The majority of these definitions are derived from INfrastructure for SPatial Information (INSPIRE), which makes the following definitions:

\begin{description}
    \item[building] ``constructions above and/or underground which are intended or used for the shelter of humans, animals, things, the production of economic goods or the delivery of services and that refer to any structure permanently constructed or erected on its site''
    \item[residential domain / residential use] ``Areas used dominantly for housing of people. The forms of housing vary significantly between, and through, residential areas. These areas include single family housing, multi-family residential, or mobile homes in cities, towns and rural districts if they are not linked to primary production. It permits high density land use and low density uses. This class also includes residential areas mixed with other non-conflicting uses and other residential areas''
\end{description}

In this paper, we follow the INSPIRE and OSM definitions of \emph{building}, and explicitly exclude temporary shelters.

\end{document}